\def\year{2021}\relax
\documentclass[letterpaper]{article} % DO NOT CHANGE THIS
\usepackage{aaai21}  % DO NOT CHANGE THIS
\usepackage{times}  % DO NOT CHANGE THIS
\usepackage{helvet} % DO NOT CHANGE THIS
\usepackage{courier}  % DO NOT CHANGE THIS
\usepackage[hyphens]{url}  % DO NOT CHANGE THIS
\usepackage{graphicx} % DO NOT CHANGE THIS
\usepackage{color,xcolor}
\usepackage{natbib}  % DO NOT CHANGE THIS AND DO NOT ADD ANY OPTIONS TO IT
\usepackage{caption} % DO NOT CHANGE THIS AND DO NOT ADD ANY OPTIONS TO IT
\usepackage{booktabs}
\usepackage{bm}
\usepackage{makecell}
\usepackage{multirow}
\usepackage{amsmath}
\usepackage{mathrsfs}
\usepackage{amssymb}
\usepackage[linesnumbered,ruled,vlined]{algorithm2e}
\usepackage{natbib}
\usepackage{hyperref}

\title{Coupled Layer-wise Graph Convolution for Transportation Demand Prediction}
\author{Junchen Ye$^1$, Leilei Sun$^{1,}$\thanks{ Corresponding Author}, Bowen Du$^1$, Yanjie Fu$^2$, Hui Xiong$^3$\\}

 \affiliations{
 	$^1$SKLSDE and BDBC Lab, Beihang University, Beijing 100083, China, \\ $^2$Department of Computer Science, University of Central Florida, FL 32816, USA, \\$^3$Department of Management Science and Information Systems, Rutgers University, NJ 07102, USA.

\{yjchen,leileisun,dubowen\}@buaa.edu.cn, yanjie.fu@ucf.edu, hxiong@rutgers.edu.
}
\begin{document}

\maketitle

\begin{abstract}

Graph Convolutional Network (GCN) has been widely applied in transportation demand prediction due to its excellent ability to capture non-Euclidean spatial dependence among station-level or regional transportation demands. However, in most of the existing research, the graph convolution was implemented on a heuristically generated adjacency matrix, which could neither reflect the real spatial relationships of stations accurately, nor capture the multi-level spatial dependence of demands adaptively. To cope with the above problems, this paper provides a novel graph convolutional network for transportation demand prediction. Firstly, a novel graph convolution architecture is proposed, which has different adjacency matrices in different layers and all the adjacency matrices are self-learned during the training process. Secondly, a layer-wise coupling mechanism is provided, which associates the upper-level adjacency matrix with the lower-level one. It also reduces the scale of parameters in our model. Lastly, a unitary network is constructed to give the final prediction result by integrating the hidden spatial states with gated recurrent unit, which could capture the multi-level spatial dependence and temporal dynamics simultaneously. Experiments have been conducted on two real-world datasets, NYC Citi Bike and NYC Taxi, and the results demonstrate the superiority of our model over the state-of-the-art ones.

% Graph neural network has been successfully applied in transportation demand prediction problem in which there are extensive topological structures. However, the adjacency matrix which determines the aggregation manner of graph neural network is fixed and mostly generated by domain knowledge. Additionally, only one adjacency matrix is used throughout the whole neural network. In this graph convolution framework, it is difficult to capture complex spatial dependencies efficiently and accurately. To bridge this gap, we propose a novel graph convolutional network. In particular, a set of layer-wise adjacency matrices are extracted by a data-driven method to capture multi-level dependencies. Then, we gradually construct high-level graphs on the basis of low-level graphs with latent relations between the topological structures in different layers, which also enables us to reduce the computational cost. Last but not least, a hierarchical aggregation automatically attaches different importance to the representation in each layer. The above components are integrated into a sequence to sequence architecture to make multi-step predictions. Experiments on two real-world datasets, NYC Citi Bike and NYC Taxi, demonstrate the superiority of our model.
\end{abstract}

\section{Introduction}

% Recent years have witnessed the rapid development and expansion of
% the cities, at the meanwhile the information technology service and Artificial Intelligence 
% As the rapid development and expansion of the cities,  

% Recent years have witnessed the rapid growth of Intelligent Transportation System (ITS), which 
% which is brought by the dramatic development and expansion of cities meeting huge development of big data on information

Recently Intelligent Transportation System (ITS) has become a hot research spot, which is mainly contributed by the two following factors: 1) the rapid development of urban transportation, and 2) the wide application of big data technology in transportation information system. Even though, there are still several important problems that remain to be dug into, such as traffic environment monitoring, travel route recommendation, in which transportation demand prediction is the most essential and crucial issue. By obtaining the precise number of the demand in each region in advance, the transport resource can be pre-allocated and re-balanced to maximize the transportation capacity, and residents will be provided with a better service in daily travel.

% Transportation demand prediction is an indispensable part of Intelligent Transportation System (ITS) and urban sensing. Recent years have witnessed the rapid growth of sharing transportation. On the one side, the on-demand vehicle (cars and bikes) sharing services bring huge convenience for daily travel of residents. On the other side, the new rising market attracting several powerful enterprises leads to a competitive circumstance. Demand prediction is a key issue to help them to seize the preemptive opportunity. By obtaining the precise number of transport demand in each region in advance, companies can pre-allocate and re-balance the transport tools to maximize the transportation capacity and provide a better service for consumers.

% Recent years have witnessed the rapid growth of sharing economy, in which sharing transportation plays a vital role. 
% The new rising market attracting several powerful enterprises leads to a competitive circumstance where demand prediction is a key issue to help them to seize the preemptive opportunity. According to obtaining the precise number of transport demand in each region in advance, companies can pre-allocate and re-balance the transport tools to maximize the transportation capacity and provide a better service for consumers.
\begin{figure}[tb!]
	\centering
	\includegraphics[width=1\columnwidth]{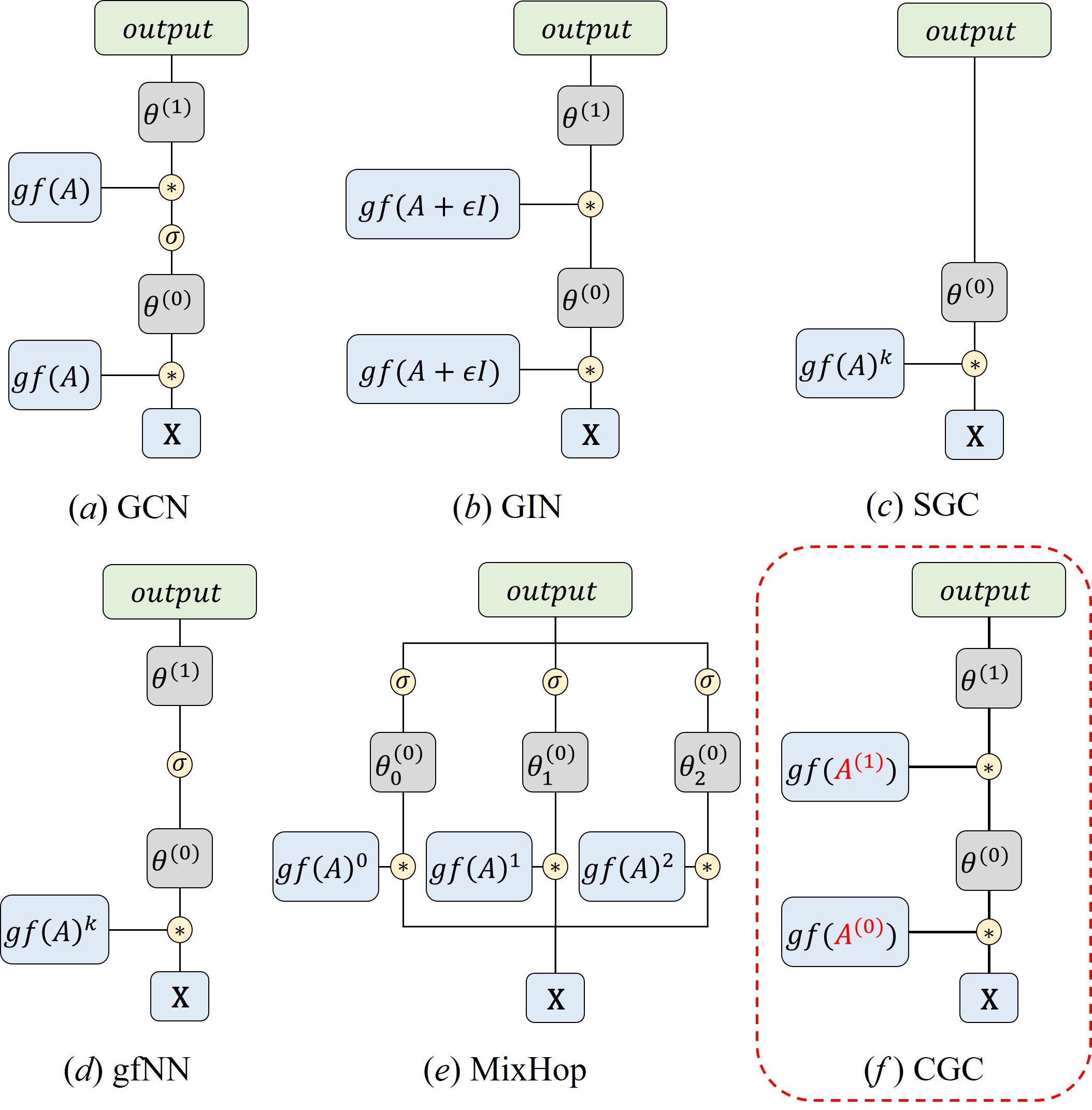}
	\caption{A visual comparison of CGC with other GCNs.}
	\label{fig:egc}
\end{figure}
% A visual comparison of CGC with other GCNs.

% traffic conditions and entangled spatial and temporal dependency,
% rove that graph convolutional network is as powerful as Weisfeiler-Lehman (WL) graph isomorphism test \cite{weisfeiler1968reduction}

% (a): GCN (Graph Convolutional
% Network) is introduce a first-order Chebyshev polynomials filter approximation. 
% (b): GIN (Graph Isomorphism Network) is proposed to prove that graph convolutional network is as powerful as Weisfeiler-Lehman (WL) graph isomorphism test with adding an $\epsilon\bm{I}$ to the adjacency matrix, where $\epsilon$ is a learnable parameter or a fixed scalar. 
% (c): SGC (Simplifying Graph Convolution) simplifies multi-layer graph convolutional network by multiplying the initial adjacency matrix by itself $k$ times.
% (d): gfNN (graph filter Neural Network) added an activation function and mapping function based on SGC to model non-linear correlations.
% (e): MixHop explore the latent representations from immediate neighbors and further neighbors with mixing powers of the adjacency matrix.

% However, most of the existing researches extract the information with the initial adjacency matrix and its high powers, which makes it difficult to capture multi-level dependencies efficiently. 
% reduce the computational complexity Graph

Due to the importance of transportation demand prediction, many efforts have been made in this field in the past few years. The development of mainstream could be divided into three stages:
1) The researches in the early time employed the empirical statistical analysis which mainly focus on forecasting the demand in a specific region instead of the whole city \cite{moreira2013tp,liu2016rebalancing}, and the incapability of capturing spatial and temporal correlation simultaneously leads to relatively low prediction accuracy. 
2) Recently, deep learning grows up quickly and gives us a new solution of modeling spatio-temporal correlations. 
% 2) Recently, quickly developing deep learning gives us a new solution of modeling spatio-temporal correlations. 
By treating the entire city as an image and partitioning it into several grids, researchers employ Convolutional Neural Network (CNN) to extract spatial correlations \cite{zhengyu2017residual} and Recurrent Neural Network (RNN) to extract temporal correlations \cite{yao2018yjp}, which makes a huge progress on the formalization. However, aggregating with neighbors in space makes CNN-based methods insensitive to long-distance transition patterns and only fit for Euclidean spatial relationship.
3) Graph convolutional network, a generalization of CNN, is fit to deal with non-Euclidean data naturally. 
Due to topological structure of road networks, it has been successfully and widely applied in transportation field. 
% Due the to topological structure of road networks, graph convolutional network (GCN), a generalization of CNN on non-Euclidean data, has been successfully and widely applied in traffic problem.
\citet{li2017dcrnn} first extracted spatial dependency with diffusion graph convolution network in traffic forecasting problem.
% \citet{guo2019attention} employed spatial-temporal attention mechanism to learn the dynamic dependency.
% \citet{wu2019wavenet} used a self-adaptive adjacency matrix to capture hidden spatial correlations.
\citet{yu2017stgcn} combined GCN with casual convolutional network, and stacked the blocks to capture spatio-temporal dependence simultaneously.

Though GCN has demonstrated its effectiveness in transportation prediction problem,
% Those researches above demonstrate the effectiveness of graph convolutional network in transport prediction problem.
there are still four important issues which have not been discussed elaborately: 
1) The adjacency matrix which determines the aggregation manner in the graph convolutional network is mostly fixed and generated by heuristic methods according to spatial distance or network connectivity, which cannot capture the genuine spatial dependence.
2) Existing methods ignore the hierarchical dependence of transportation demand prediction. For example, a sudden rainstorm causes the global reduction of sharing bike usage, but a congestion led by traffic accident can only make a local impact.
3) Mainly following the perspective of graph signal processing, current graph convolution approaches tend to smooth nodes' input signals. In this situation, it is difficult for stacked graph convolution layers with only one adjacency matrix to obtain multi-level representations of transportation demand efficiently.
4) The representations in different layers contributing to the final transportation demand should not be static but dynamic over time. For example, a traffic emergency may raise the influence of low-level features. 
Obviously, the above four problems could be exploited to improve current demand prediction research, which has been rarely discussed in existing research.
% from a new perspective.
% To address aforementioned challenges, we make efforts in transport prediction and adjacency matrix generation two aspects.
% To capture multi-level representations and model temporal dynamics, a novel deep learning framework named \textit{\textbf{E}volution \textbf{C}onvolutional \textbf{R}ecurrent \textbf{N}eural \textbf{N}etwork} (ECRNN) is proposed. First, we design \textit{\textbf{E}volution \textbf{G}raph \textbf{C}onvolution} (EGC) to overcome the difficulty of multi-level representation extracting. The particular graph convolution network possesses evolutionary and adaptive adjacency matrices in stacked layers. Then, we propose a hierarchical aggregation method to fuse the multi-level information. Last, we build the temporal prediction with the sequence to sequence architecture. We also propose an external factor information free method to generate adjacency matrix. The totally data-driven approach captures the similarity of nodes' extracted features. In summary, this paper has the following contributions:

To cope with aforementioned issues, a novel deep learning framework named \textit{\textbf{C}oupled Layer-wise \textbf{C}onvolutional \textbf{R}ecurrent \textbf{N}eural \textbf{N}etwork} (CCRNN) is proposed.
% First, a novel graph convolution structure with adaptive adjacency matrices varying from layer to layer is proposed to extract multi-level dependencies efficiently and accurately, namely, \textit{\textbf{C}oupled Layer-wise \textbf{G}raph \textbf{C}onvolution} (CGC). 
% , namely, \textit{\textbf{C}oupled Layer-wise \textbf{G}raph \textbf{C}onvolution} (CGC).
% First, we provide a novel graph convolution structure which has different adjacency matrices in different layers. The genuine topological structure is extracted by self-learned adjacency matrices.
% Figure \ref{fig:egc} shows a visual comparison between CGC and several popular graph convolutional networks.
% Second, a layer-wise coupled mapping mechanism is proposed to build the upper-level graph upon on the lower-level one with latent correlations between topological structure in different layers.
% Last, a multi-level aggregation automatically attaches the importance to the representation in each layer. Above components are integrated into a sequence to sequence architecture to make multi-step predictions.
% Last, an domain knowledge free method is employed to generate adjacency matrix based on the similarity of nodes' temporal patterns. 
% Last, an external factor information free method is employed to generate adjacency matrix. The totally data-driven approach captures the similarity of nodes' extracted features.
Specifically, to address the problem that it is difficult for the popular GCNs to capture multi-level spatial dependence efficiently and accurately, we propose a novel graph convolution architecture, \textit{\textbf{C}oupled Layer-wise \textbf{G}raph \textbf{C}onvolution} (CGC), with self-learned adjacency matrices varying from layer to layer. Figure \ref{fig:egc} shows a visual comparison between CGC and several popular graph convolutional networks. Furthermore, by modeling the layer-wise topological structure correlations, we provide a coupled mapping mechanism to implement this graph convolution structure at a small computational cost. The extracted representations are attached different importance by a multi-level aggregation module. Finally, the above components are fused with a recurrent unit to capture spatio-temporal correlations simultaneously.
In summary, this paper has the following contributions:
\begin{itemize}
	\item \textit{A novel graph convolution architecture} is proposed to extract multi-level spatial dependence adaptively. This structure has different adjacency matrices in different layers and all adjacency matrices are self-learned during the training process.
    % \item \textit{A layer-wise coupling mechanism} is proposed to bridge the adjacency matrices construction mechanism by coupled  mapping. The upper-level adjacency matrix is built upon on the lower-level one according to the latent correlations between topological structure in different layers. It also reduces the scale of parameters in our model.
    \item \textit{A layer-wise coupling mechanism} is proposed to bridge the upper-level adjacency matrix with the lower-level one according to the hidden correlations of topological structures in different layers. 
    It also reduces computational costs during the training process.
    % It also reduces the scale of parameters in our model.
    % A decomposing and integrating adjacency matrices method is leveraged to reduce the computational cost.
    % \item A hierarchical aggregation method is proposed to fuse multi-level representations. Based on this, we proposed HCCRNN, \textit{an unitary prediction framework} integrating the spatial hidden states with Gated Gate Recurrent Unit (GRU) in a sequence to sequence architecture.
    % \item \textit{An unitary prediction framework, \textbf{C}oupled Layer-wise \textbf{C}onvolutional \textbf{R}ecurrent \textbf{N}eural \textbf{N}etwork} (CCRNN), is proposed to integrate the spatial hidden states with Gated Gate Recurrent Unit (GRU) in a sequence to sequence architecture. The representations is fused by a multi-level aggregation.
    \item \textit{A unitary prediction framework} is proposed to make the final prediction by integrating the spatial hidden states with gated recurrent unit in a sequence to sequence architecture, where the spatial hidden states are obtained by aggregating multi-level demand dependence.
    % is proposed to make the final prediction by integrating the spatio-temporal hidden states with gated recurrent unit in a sequence to sequence architecture
\end{itemize}

\section{Related Work}
In this section, we review the literature related to our work 
from the perspectives 
of transportation demand prediction and graph convolution.
\subsection{Transportation Demand Prediction} 
Huge efforts have been made in traffic prediction due to the years of continuous research. In the early time, the attention was mainly paid to combining data mining methods and empirical statistical analysis \cite{moreira2013tp, liu2016rebalancing, tong2017simpler}. 
Those methods have limited researched region and spatio-temporal correlations could not be captured simultaneously. Deep learning methods provide a new perspective to deal with non-linear relations.
\citet{lv2014sae} used a SAE (stacking autoencoders) to extract representations.
\citet{zhengyu2017residual} treated the demands in the whole city as an image at each time step, and leveraged popular models in CV field at that time, residual network, to capture correlations between different grids.
Following this line, \citet{yao2018yjp} proposed a multi-view network incorporating CNN and Long Short-Term Memory (LSTM) to capture sptatio-temporal correlations simultaneously. 
\citet{zhou2018cluster} clustered the demand snapshots to select representations.
\citet{ye2019coprediction} explored the correlations between different modes of transportation and made a co-prediction. However, it is hard for CNN-based methods to capture long-distance transition patterns, which makes graph convolutional network step onto the stage \cite{yu2017stgcn, bai2019stg2seq, yu2020deep}.
\citet{li2017dcrnn} unified diffusion convolutional layer with Gated Recurrent Unit (GRU) in an encoder-decoder architecture.
\citet{chen2019bicomponent} explored the correlations of both nodes and edges with two types adjacency matrices.
\citet{yu20193d} proposed a 3D graph convolutional network.
It is worth mentioning that \citet{wu2019wavenet} employed an random initialized adaptive adjacency matrix to capture the hidden spatial dependency precisely. However, it is difficult for grid-based data incorporating with graph convolution based methods to get satisfying results.

\subsection{Graph Convolutional Network}
Graph convolutional network which generalizes convolutional neural network to non-Euclidean data has received increasing attention over past few years. \citet{bruna2013spectral} extended the convolution network to graph-based data. \citet{chebnet2016} reduced the computational cost with Chebyshev polynomials. A first-order Chebyshev polynomials approximation was introduced by \citet{kipf2016gcn}. 

Graph convolutional networks fall into two main categories, spatial-based and spectral-based \cite{wu2020survey}.
Spatial-based methods update information by designing different strategies to aggregate features of their neighbors \cite{hamilton2017graphsage}.
\citet{velivckovic2017gat} employed attention mechanisms to learn weights between two nodes.
Spectral-based methods treat graph convolution operation as removing noises from graph signal, and the key to this stream is the structure and capacity of the filter. 
Along this line, 
% \citet{xu2018howpowerful} proved the expressiveness of graph neural networks is equal to Weisfeiler-Lehman (WL) graph isomorphism test.
\citet{chiang2019clustergcn} weakened the influence of neighbors by adding an identity matrix to adjacency matrix. 
% \citet{wu2019simplify} multiplied the adjacency matrix itself to simplify the multi-layer GNN structure.
\citet{xu2019heat} enhanced low-frequency filters with the heat kernel.
\citet{li2018adaptive} learned an adaptive residual Laplacian matrix with generalized Mahalanobis distance.
However, they all leverage the initial adjacency matrix and its variants, which fails to capture multi-level dependence efficiently and accurately.

% crucial information is hidden in the arrival and departure records.
\section{Preliminaries}
% In this section, we introduce the details of CCRNN. shows the architecture of our proposed method.
In this section, we present several definitions and the problem formalization.
\subsection{Station-level Demand Prediction}
\textbf{Transportation Station.} 
The modes of transportation could be divided into two categories, station-based and station-less. 
For the station-based transportation, such as bus and subway, it is intuitive to be formalized as graph structure by denoting each station as a node. 
For the station-less transportation, such as taxi and sharing bike, although the locations of passengers' arrival and departure are discrete, they tend to gather around certain places. For example, there are many taxi order requests at the main entrance of a university, which forms a virtual station naturally \cite{huxiao2020gnn}. Discovering potential stations can help to capture transportation demand features and make predictions more accurately.
It is worth mentioning that most recent deep learning based demand prediction methods partition the city into grids to meet the requirement of CNN. We employ a Density Peak Clustering (DPC) \cite{huxiao2014dpc} based method to discover virtual stations. 

% For the  which are represented as 
% The real stations and discovered stations
% $\mathbb{S}= \{s_1, s_2,...,s_n,...,s_N \}$ where $N$ is the total number of stations.

% In the dockless transportation system, the virtual stations $\mathbb{S}$ correspond to the nodes $V$ in the study region.
\textbf{Adjacency Matrix.} 
Given a graph $G = (V,E)$, we denote $V$ as the set of nodes and $E$ as the set of edges. In the transportation system, the stations correspond to the nodes $V$ in study regions.
At time step $t$, the graph $G$ has a feature matrix $\mathbf{X}_t \in \mathbf{R}^{N\times d}$ where $d$ is the input feature dimension and $N$ is the number of nodes. Given graph signals of $\tau$ time steps, we aim at obtaining an adjacency matrix $\bm{A}\in\mathbf{R}^{N\times N}$ with a data-driven method to complete the definition of $G$. The mapping function $\mathcal{F}_1$ is described as:
\begin{equation}
[\mathbf{X}_{t_{a}:t_{a}+\tau-1}] \xrightarrow{\mathcal{F}_1} \bm{A},
\end{equation}
where $t_{a}$ denotes the first time step in this adjacency matrix generating problem.

\textbf{Transportation Demand Prediction.} 
At time step $t$, given the graph $G$ and $P$ steps historical graph signals, we intend to obtain a mapping function $\mathcal{F}_2$ to forecast next $Q$ steps graph signals. This can be defined as:
\begin{equation}
[\mathbf{X}_{t-P+1:t} ,G] \xrightarrow{\mathcal{F}_2} \mathbf{X}_{t+1:t+Q},
\end{equation}
where $\mathbf{X}_{t+1:t+Q} \in \mathbf{R}^{Q\times N \times d}$ and $\mathbf{X}_{t-P+1:t} \in \mathbf{R}^{P\times N \times d}$.

\subsection{Graph Convolutional Network}
% \textbf{Graph Convolution Unit.}
Given a graph $G = (V,E)$, we denote $\hat{\bm{A}}$ as the normalized adjacency matrix:
\begin{equation}
\label{equation:randomwalk}
\hat{\bm{A}} = \bm{D}^{-1}\bm{A},
\end{equation}
where $\bm{A}$ is the adjacency matrix, $\bm{D}$ is a diagonal matrix of node degrees, $\bm{D}_{ii} = \sum_j\bm{A}_{ij}$. Following \cite{li2017dcrnn},
we remove the activation function and model the diffusion process with $K$ steps under undirected graph structure,
which leads to the final feature propagation equation:
\begin{equation}
\bm{X} \star_{G} \bm{g_\theta} = \sum_{i=0}^{K} (\hat{\bm{A}})^i\bm{X}\bm{\theta_i},
\end{equation}
where $\bm{g_\theta}$ is the filter with the parameters $\bm{\theta}$, and $\bm{X}$ is the input signals.

\textbf{An Unifying View of the Existing GCNs.}
Spectral-based graph convolutional network research mainly focuses on the definition of convolution filter $\bm{g_\theta}$. As illustrated in Figure \ref{fig:egc}, 
(\textit{a}) GCN introduces a first-order Chebyshev polynomials filter approximation \cite{kipf2016gcn}, which has been widely adopted;
(\textit{b}) GIN (Graph Isomorphism Network) is constructed with an added weighted identity matrix on the adjacency matrix \cite{xu2018howpowerful}. They proved graph convolutional network is as powerful as Weisfeiler-Lehman (WL) graph isomorphism test \cite{weisfeiler1968reduction};
(\textit{c}) SGC (Simple Graph Convolution) simplifies the multi-layer graph convolutional network by multiplying the initial adjacency matrix by itself $k$ times \cite{wu2019simplify};
(\textit{d}) gfNN (graph filter Neural Network) adds an activation function and a mapping function on the basis of SGC in order to model non-linear correlations \cite{nt2019gfnn};
(\textit{e}) MixHop explores the latent representations of the immediate neighbors and further neighbors by mixed powers of the adjacency matrix \cite{abu2019mixhop}.

Those researches employed graph convolution with the initial adjacency matrix and its high powers, which makes it difficult to capture multi-level dependence efficiently. As shown in Table \ref{fig:egc} (\textit{f}), to address this problem, CGC extracts the hierarchical representations with self-learned adjacency matrices varying from layer to layer.

\section{Methodology}
\begin{figure*}[tbh!]
	\centering
	\includegraphics[width=2\columnwidth]{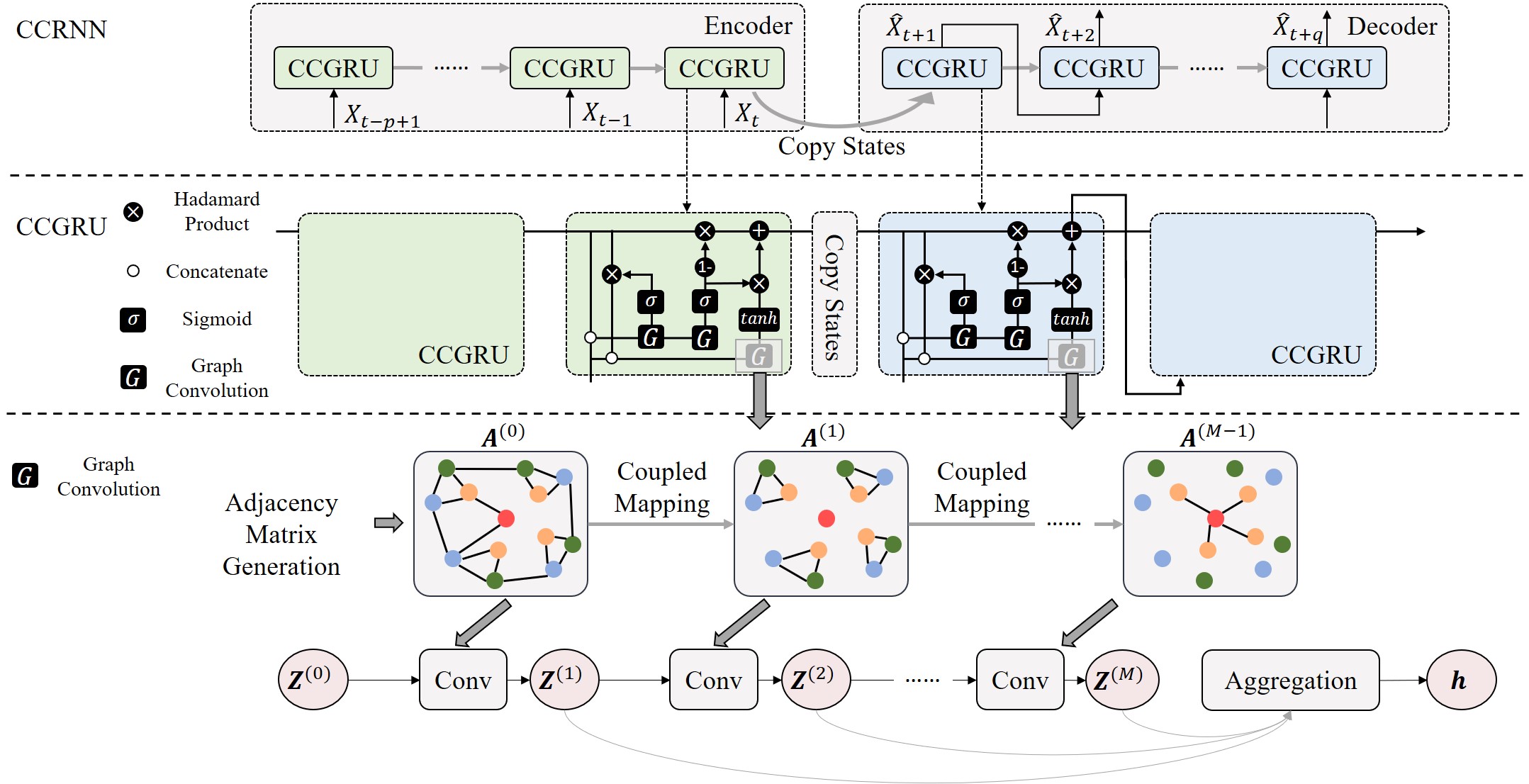}
	\caption{The framework of CCRNN.}
	\label{fig:model}
\end{figure*}

% In this section, we detail the structure of ECRNN.
% Firstly, the method of adjacency matrix construction will be described.
% Next, we elaborate the structure of Evolution Graph Convolution (EGC). A hierarchical aggregation is employed to learn the weights of representations in different layers.
% Finally, for modeling the temporal dynamics, 
% we fuse the feature extracting components above with Gate Recurrent Unit (GRU) to form an unified framework. Figure \ref{fig:model} shows the architecture of our proposed method (ECRNN).
In this section, we introduce the details of CCRNN. Figure \ref{fig:model} shows the architecture of our proposed method.

\subsection{Adjacency Matrix Generation}
Adjacency matrix is crucial as it determines the aggregation manner of nodes themselves and their neighbors in graph convolution.
Different from the proposed methods which use hand-designed features to construct the graph structure in advance, our adjacency matrix generation method is totally data-driven, and temporal correlations are extracted.
% will be extracted and it still works without external factors.

Given a set of graph signals $\mathbf{X}_{t_{a}:t_{a}+\tau-1}\in\mathbf{R}^{\tau \times N \times d}$, 
we reshape this 3-D tensor as a 2-D matrix shaped $(\tau \cdot d) \times N$. To capture interior similarity between different stations and filter the redundant information among stations, we decompose the 2-D matrix $\bm{X}^a$ into two:
\begin{equation}
    \bm{X}^a = \bm{X}^t {\bm{X}^s}^T,
\end{equation}
where $\bm{X}^t,\bm{X}^s$ indicate the time-wise and station-wise matrices. Actually, we use Singular Value Decomposition (SVD) to decompose $\bm{X}^a$ in experiments.
% Actually, We decompose the matrix by Singular Value Decomposition (SVD) and multiply the singular value matrix with one of the unitary matrix to get $\bm{X}^t,\bm{X}^s$.

% In the classical recommendation system with collaborative filtering methods, the preference matrix is decomposed into a user-based matrix and an item-based matrix. To better understand our propose, we have an analogy between user-based matrix and station-wise matrix. 
A large number of repeated transportation patterns are hidden in $\bm{X}^a$. SVD could filter the redundant information out by dimension reduction. $\bm{X}^s \in \mathbf{R}^{N \times \xi}$ contains the compact and high-level representations of each station, where $\xi$ is the dimension of the stations' features. We calculate the similarity of $x$-th and $y$-th rows of $\bm{X}^s$ as their edge weight in adjacency matrix:
\begin{equation}
\label{equation:similarity}
    \bm{A}_{xy} = Similarity(\bm{X}_x^s, \bm{X}_y^s),
\end{equation}
where $\bm{A}_{xy}$ indicates the $x$-th row and $y$-th column of the adjacency matrix $\bm{A}$. Here we use the Gaussian kernel based method to estimate the pairwise similarity. The Equation (\ref{equation:similarity}) can be redefined as:
\begin{equation}
    \bm{A}_{xy} = exp(-\frac{\|\bm{X}_x^s-\bm{X}_y^s\|^2}{\varepsilon^2}),
\end{equation}
where $\varepsilon$ is the standard deviation.

\subsection{Coupled Layer-wise Graph Convolution}
Previous works partition the city into regular grids to meet the requirement of CNN, but the fixed and local receptive field of CNN makes it hard to capture the features of long-distance transition and the similarity between regions. 
The theory of spectral graph generalizes the convolution operation from regular gird structure to graph structure. However, it is still unwise to use graph convolution on gird-based data to extract high-level representations of each time step to make precise predictions, because there might be multiple modes of transportation patterns mixed together in one gird simultaneously.
% To address those problems, we introduce a unify definition of transportation station to make a unified graph formalization for station-based and station-less transportations.
To address this problem, we introduce a unified graph formalization for station-based and station-less transportations.

To capture multi-level dependence efficiently and accurately, we propose a novel graph convolutional network, Coupled Layer-wise Graph Convolution (CGC), which has different adjacency matrices in different layers. This structure can be defined recursively:
\begin{equation}
\label{equation:initEGC}
    \bm{Z}^{(m+1)} = \bm{Z}^{(m)} \star_{G} \bm{g_\theta}^{(m)} = \sum^K_{i=0} (\bm{A}^{(m)})^i\bm{Z}^{(m)}\bm{\theta}_i^{(m)},
\end{equation}
where $\bm{Z}^{(m)}$ represents the input of layer $m+1$. $\bm{Z}^{(m+1)}$ is not only the output of layer $m+1$ but also the input of layer $m+2$. The multi-level relationship of nodes is modeled by $\bm{A}^{(m)}$ which varies from layer to layer. We use a coupled mapping function $\psi^{(m)}$ to construct upper-level adjacency matrix in layer $m+1$:
\begin{equation}
    \bm{A}^{(m+1)} = \psi^{(m)}(\bm{A}^{(m)}).
\end{equation}

%但是会随机初始化的矩阵带来训练困难，梯度选择问题，无法收敛到最好,同时无法与天然引入的外部矩阵结合？？？（引入外部矩阵，无法学习统一的邻接关系表示）。所以我们提出一种新的框架，我们使用xxx作为initial，在此基础上进行自适应with梯度优化。firstlayer i
Self-adaptive convolution multiplies two random initial matrices which does not require any prior knowledge and matrices are trained end-to-end through stochastic gradient descent \cite{wu2019wavenet}. In this way, a new adjacency matrix is generated to extract hidden spatial dependencies. However, random initialization brings the difficulty of convergence and numerical instability.
To solve this problem, we initialize the self-adaptive matrix with original graph structure and employ stochastic gradient descent to optimize it. The first layer of CGC is defined as:
\begin{equation}
\label{equation:egcfirstlayer1}
    \bm{Z}^{(1)} =  \sum^K_{i=0}(\bm{A}^{(0)})^i\bm{Z}^{(0)}\bm{\theta}_i^{(0)},
\end{equation}
where $\bm{Z}^{(0)} = \bm{X}$. The feature matrix $\bm{X} \in \mathbf{R}^{N\times d}$ at each time step is fed into the first layer of CGC as input. In this paper, 
% $\bm{A^0}$ is initialized by the random walk normalized adjacency matrix generated by Equation \ref{equation:similarity}
we employ Equation (\ref{equation:randomwalk}) to normalize the adjacency matrix generated by Equation (\ref{equation:similarity}).
The output is taken as original $\bm{A}^{(0)}$, and we optimize it with stochastic gradient descent to discover the real spatial relationships.

% Additionally, we feed the matrix generated from Section of Adjacency Matrix Generation Problem into Equation \ref{equation:randomwalk}, and take the output as $\bm{A^0}$ which is set learnable in training process.

% Additionally, we normalize the adjacency matrix generated from Equation \ref{equation:similarity} with Equation \ref{equation:randomwalk}, and take the 

% The self-adaptive matrix is initialized by normalized adjacency matrix 

% above formulation leads to over parameterization with high costs of $N \times N$ adjacency matrix and $\psi$ caused by the huge number of nodes. 
% However, due to the huge number of nodes, the high computational costs of $N \times N$ adjacency matrix and $\psi$ lead to the over-parameterization in above formulations.
However, due to the huge number of nodes, $N \times N$ adjacency matrix and $\psi$ obtain high computational costs, which lead to the over-parameterization in above formulations.
To solve this problem, we decompose $\bm{A}^{(0)} \in \mathbf{R}^{N \times N}$ into two small matrices with SVD:
\begin{equation}
    \bm{A}^{(0)} = \bm{E}_1^{(0)}{\bm{E}_2^{(0)}}^T,
\end{equation}
where $\bm{E}_1^{(0)} \in \mathbf{R}^{N \times L}$ is the source node embedding in the first layer, $\bm{E}_2^{(0)} \in \mathbf{R}^{N \times L}$ is the target node embedding, and $L$ is the representation dimension. The number of trainable parameters is reduced from $N \times N$ to $2 \times N \times L$ ($N>>L$). Equation (\ref{equation:egcfirstlayer1}) can be redefined as:
\begin{equation}
    \bm{Z}^{(1)} =  \sum^K_{i=0}(\bm{E}_1^{(0)}{\bm{E}_2^{(0)}}^T)^i\bm{Z}^{(0)}\bm{\theta}_i^{(0)}.
\end{equation}
Additionally, function $\psi$ is  implemented by fully connected layer to model the layer-wise correlations. We share the parameters of fully connected layers between $\bm{E}_1$ and $\bm{E}_2$:
\begin{equation}
\begin{aligned}
\label{equation:evolutionconv}
    \bm{E}^{(m)}_1 =& \bm{E}^{(m-1)}_1\bm{W}^{(m-1)} + \bm{b}^{(m-1)}, \\
    \bm{E}^{(m)}_2 =& \bm{E}^{(m-1)}_2\bm{W}^{(m-1)} + \bm{b}^{(m-1)},  \\
\end{aligned}
\end{equation}
where $\bm{E}^{(m)}_1$ and $\bm{E}^{(m)}_2$ denote the source node embedding and target node embedding in $m$-th layer. $\bm{W}^{(m-1)}$ and $\bm{b}^{(m-1)}$ represent the weight and bias. To simplify the model, the feature dimensions of $\bm{E}_1$ and $\bm{E}_2$ in different layers are all set as $L$. 
% We reduce the the parameter number of each coupled mapping function $\psi$ to $L \times (L+1)$.
The parameter number of each coupled mapping is reduced to $L \times (L+1)$.

According to Equation (\ref{equation:initEGC}) and (\ref{equation:evolutionconv}), the mathematical expression of CGC can be redefined recursively as:
% \begin{equation}
% \label{equation:finalEGC}
% \begin{aligned}
%      \bm{Z}^{(m+1)} = &\bm{Z}^{(m)} \star_{G} \bm{g_\theta}^{(m)} \\
%      = &\sum^K_{i=0}\theta_i^{(m)} (\bm{E}_1^{(m)}{\bm{E}_2^{(m)}}^T)^i\bm{Z}^{(m)}
% \end{aligned}
% \end{equation}
\begin{equation}
\label{equation:finalEGC}
     \bm{Z}^{(m+1)} = \sum^K_{i=0} (\bm{E}_1^{(m)}{\bm{E}_2^{(m)}}^T)^i\bm{Z}^{(m)}\bm{\theta}_i^{(m)}.
\end{equation}
It is worth mentioning that only $\bm{E}_1,\bm{E}_2$ in the first layer are optimized straightforward by stochastic gradient descent, and the others are updated by coupled mapping $\psi$.

%Different from self-adaptive adjacency matrix generating to address the problem of high-level representation extracting and promote the stability and accuracy of self-adaptive convolution. 

\subsection{Multi-level Aggregation}
%讲attention的好处，到一点发展
For gathering the information from all graph convolution layers rather than extracting from only one fixed layer, we implement the multi-level aggregation by an attention mechanism to select the information which is relatively important to the current prediction task.

% We employ a linear transformation to calculate the attention scores
% The attention scores are calculated by a linear transformation  attention mechanism
Graph signals' multi-level representations obtained by CGC are denoted as $\mathbb{Z} = \{\bm{Z}^{(1)},\bm{Z}^{(2)},...,\bm{Z}^{(m)},...,\bm{Z}^{(M)}\}$, $\mathbb{Z} \in \mathbf{R}^{M \times N \times \beta}$, where $M$ represents the total number of graph convolution layers, and $\beta$ denotes the dimension of features. The attention scores are calculated by a linear transformation,
and Softmax function helps to normalize the coefficients. With summing up the outputs of multiplying the $\mathbb{Z}$ and normalized scores, the aggregation is defined as:
\begin{equation}
   \alpha^{(m)} =  \frac{exp(\bm{\hat{Z}}^{(m)}\bm{W}_\alpha+b_\alpha)}{\sum_{m=1}^{M} exp(\bm{\hat{Z}}^{(m)}\bm{W}_\alpha+b_\alpha)},
\end{equation}
\begin{equation}
   \bm{h} = \sum^M_{m=1} \alpha^{(m)} \bm{Z}^{(m)},
\end{equation}
where $\bm{W}_\alpha$ and $b_\alpha$ represent the weight and bias in linear transformation, $\bm{\hat{Z}}^{(m)}$ is the flattened version of $\bm{Z}^{(m)}$, and $\alpha^{(m)}$ is the attention score of $\bm{Z}^{(m)}$. $\bm{h}$ is the final result of CGC, and the output will be fed into GRU.

\subsection{Temporal Dependence Modeling}
GRU, a simple but powerful variant of RNN, solves the problem of exploding and vanishing gradient. Following \cite{li2017dcrnn}, we replace the linear transformation in GRU with the combination of CGC and multi-level aggregation. The Coupled Layer-wise Convolutional Recurrent Gated Recurrent Unit (CCGRU) is defined as:
\begin{equation}
\begin{aligned}
\bm{r}^{(t)} & =  \sigma(\Theta_{r}\star_ G[\bm{h}^{(t)},\bm{H}^{(t-1)}]+\bm{b}_r), \\
\bm{u}^{(t)} & =  \sigma(\Theta_{u}\star_ G[\bm{h}^{(t)},\bm{H}^{(t-1)}]+\bm{b}_u), \\
\bm{c}^{(t)} & =  tanh(\Theta_{c}\star_ G[\bm{h}^{(t)},(\bm{r}^{(t)}\odot\bm{H}^{(t-1)})] + \bm{b}_c), \\
\bm{H}^{(t)} & = \bm{u}^{(t)} \odot \bm{H}^{(t-1)} + (1-\bm{u}^{(t)}) \odot \bm{c}^{(t)},
\end{aligned}
\end{equation}
where $\bm{h}^{(t)}$ and $\bm{H}^{(t)}$ represent the result of attention mechanism and output of GRU at $t$ time step.
$\odot$ is the Hadamard product, and $\sigma$ is the  Sigmoid function.
Reset gate $\bm{r}^{(t)}$ helps to forget dispensable information, and the update gate $\bm{u}^{(t)}$ controls the output of GRU at time step $t$.
We denote the convolution operation in Equation (\ref{equation:finalEGC}) as $\star_G$, and $\Theta_r,\Theta_u,\Theta_c$ represent the parameters of corresponding filters. 

In multi-step forecasting model, we employ the sequence to sequence architecture and scheduled sampling strategy \cite{li2017dcrnn}.
During the training period, the final state of the encoder is copied to initialize the decoder, and the decoder obtains previous ground truth in a decaying probability to predict. The forecasting results replace the ground truth observations in the testing step. Along with the last piece of the puzzle, we finish building Coupled Layer-wise Convolutional Recurrent Neural Network (CCRNN).

\section{Experiments}
\begin{table*}[t]           % 表格
	\centering
	\caption{Evaluations of CCRNN and baselines on NYC Citi Bike and NYC Taxi.}
	\label{tab:baselinescomp}
	\begin{tabular}{cccc|ccc}
		\hline
		\multirow{2}{*}{Method} & \multicolumn{3}{c}{NYC Citi Bike} & \multicolumn{3}{c}{NYC Taxi} \\
		\cline{2-7}
		& RMSE & MAE & PCC & RMSE & MAE & PCC \\
		\hline
		\hline
		HA & 5.2003 &3.4617  & 0.1669  &29.7806 &16.1509 &0.6339 \\
		XGBoost \cite{chen2016xgboost} & 4.0494 &2.4690 & 0.4861  & 21.1994 &11.6806&   0.8077\\
		FC-LSTM \cite{hochreiter1997lstm} & 3.8139 & 2.3026 &0.5675  & 18.0708&10.2200 &0.8645 \\
		DCRNN \cite{li2017dcrnn}& 3.2094& 1.8954&0.7227
		& 14.7926 &8.4274&	0.9122	\\
		STGCN \cite{yu2017stgcn}&3.6042&	2.7605&	0.7316
		&22.6489 &	18.4551 &	0.9156\\
		STG2Seq \cite{bai2019stg2seq}&3.9843 & 2.4976&	0.5152    & 18.0450 &	9.9415 &	0.8650 \\
		Graph WaveNet \cite{wu2019wavenet}& 3.2943& 1.9911	& 0.7003&  13.0729&8.1037&	0.9322\\
		\textbf{CCRNN} & \textbf{2.8382} & \textbf{1.7404} & \textbf{0.7934}  &\textbf{9.5631} &\textbf{5.4979} &  \textbf{0.9648}\\
		\hline
	\end{tabular}
\end{table*}
% To validate the strength of our model, we conduct extensive experiments on real-world datasets. In this section, firstly, we introduce the datasets and baselines. Then experiment settings are stated elaborately. Finally, we present the detailed experiment results and provide further discussion.
In this section, experimental results and detailed analysis will be presented. The source code is available \footnote{\url{https://github.com/Essaim/CGCDemandPrediction}}.

% \url{https://github.com/Essaim/CGCDemandPrediction}
% To validate the strength of our model, we conduct extensive experiments on real-world datasets. Firstly, we introduce the datasets and list baselines. Finally, we present the detailed experiment results and provide further discussion.

\subsection{Datasets}
% Experiments are conducted on two real-world datasets collected from NYC OpenData. The two datasets contain order records of taxi and bike in NYC.
% \begin{itemize}
% 	\item \textbf{NYC Citi Bike}\footnote{https://www.citibikenyc.com/system-data}: This dataset includes the orders generated from people using NYC Citi bike in daily life. We choose the transaction records from April $1st$, 2016 to June $30th$, 2016 (91 days) which are about 3.7 million totally. Following information is contained: bike pick-up station, bike drop-off station, bike pick-up time, bike drop-off time, trip duration.
% 	\item \textbf{NYC Taxi}\footnote{https://www1.nyc.gov/site/tlc/about/tlc-trip-record-data.page}: This dataset consists of 35 million taxicab trip records in New York from April $1st$, 2016 to June $30th$, 2016 with about 380 thousand order records per day. Following information is contained: pick-up time, drop-off time, pick-up longitude, pick-up latitude, drop-off longitude, drop-off latitude, trip distance.
% \end{itemize}
Experiments are conducted on two real-world 
datasets collected from NYC OpenData. The two datasets contain order records of taxi and bike in NYC.
\begin{itemize}
	\item \textbf{NYC Citi Bike}\footnote{\url{https://www.citibikenyc.com/system-data}}: This dataset includes the NYC Citi bike orders of people daily using. We choose the transaction records from April $1st$, 2016 to June $30th$, 2016 (91 days). Following information is contained: bike pick-up station, bike drop-off station, bike pick-up time, bike drop-off time, trip duration.
	\item \textbf{NYC Taxi}\footnote{\url{https://www1.nyc.gov/site/tlc/about/tlc-trip-record-data.page}}: This dataset consists of 35 million taxicab trip records in New York from April $1st$, 2016 to June $30th$, 2016. Following information is contained: pick-up time, drop-off time, pick-up longitude, pick-up latitude, drop-off longitude, drop-off latitude, trip distance.
\end{itemize}

\subsection{Baselines}
Following methods are compared and we tune the key hyper-parameters to make sure that they have the best performance.
\begin{itemize}
    \item \textbf{HA}: We calculate the average of historical values at previous time steps as history average.
    \item \textbf{XGBoost}: XGBoost \cite{chen2016xgboost} is a widely used method based on gradient boosting tree.
    \item \textbf{FC-LSTM}: LSTM \cite{hochreiter1997lstm} incorporates with fully connected network.
    \item \textbf{DCRNN}: Diffusion convolution recurrent neural network \cite{li2017dcrnn} combines diffusion graph convolution with GRU in an encoder-decoder manner.
    \item \textbf{STGCN}: Spatial-temporal graph convolution network \cite{yu2017stgcn} combines graph convolution with casual convolution.
    \item \textbf{STG2Seq}: Spatial-temporal graph to sequence model \cite{bai2019stg2seq} can capture the long-term and the short-term information.
    \item \textbf{Graph WaveNet}: Graph WaveNet \cite{wu2019wavenet} conducts graph convolution with adaptive adjacency matrix.
\end{itemize}

\subsection{Experimental Setup}
The researched region is an $8.42km\times14.45km$ rectangle covering West New York, Manhattan Island and part of Brooklyn.
NYC Citi Bike is dock-based and every depot of bikes is considered as a station. We filter out the stations with fewer orders and keep the 250 stations with the most orders. As for dockless NYC Taxi, their orders are clustered into 266 virtual stations.
The time step length is set to half an hour, such as $0:00am$ to $0:30am$, $0:30am$ to $1:00am$, $1:00am$ to $1:30am$.
Among the last four weeks, the first two are used for validation, and the last two are for testing. 

%总长4355 3011 672 672    
The demands in all stations are standardized. The feature dimension $D$ is 2, representing the pick-up demand and drop-off demand. The historical demand length $P$ is set to 12 and the prediction length $Q$ is 12, too. 
In the adjacency matrix generation, to avoid the influence of validation and testing data, we employ the entire training dataset to learn stations' representations. That is to say, the first time step of generating adjacency matrix $t_a$ is 0, and the length $\tau$ is 3,011 (the length of training set). The dimension of station feature $\xi$ is set to 20. In CGC, the number of stacked convolution layers $M$ is 3. We use Equation (\ref{equation:finalEGC}) as final convolution layer with diffusion steps $K=3$. The dimension of two adaptive matrices $L$ is 50. The hidden states dimension $\beta$ is set to 25. Learning rates for NYC Citi Bike and NYC Taxi datasets are 0.0005 and 0.0015. For training stability, we initialize the weight $\bm{W}$ as identity matrix and bias $\bm{b}$ as 0 in coupled mapping.
All methods are optimized by Adam algorithm \cite{kingma2014adam}.
The model is implemented with the PyTorch framework. 
We choose the following three evaluation metrics: Root Mean Squared Error (RMSE), Mean Absolute Error (MAE), and Pearson Correlation Coefficient (PCC). 
RMSE between the output and the ground truth is used as the loss function.

% Experiments are conducted on an Ubuntu machine equipped with two Intel(R) Xeon(R) CPU E5-2667 v4 @ 3.20GHz with 8 physical cores, and the GPU is NVIDIA TITAN Xp, armed with 12 GB of GDDR5X memory running at over 11 Gbps.

\subsection{Main Results}
% \paragraph{Comparison with Baselines:} 
\textbf{Comparison with Baselines.}
Table \ref{tab:baselinescomp} shows the comparison results with different baselines. We evaluate models on the multi-step outputs to obtain a global view. 
Each baseline is trained for 10 times to obtain an average result. Compared with the performances on NYC Taxi, methods have relatively smaller RMSE in NYC Citi Bike since NYC Taxi has a larger demand number. Our method achieves the best performance in all evaluation metrics on two datasets. 

Poor performances of HA, XGBoost, FC-LSTM indicate the limitation of employing temporal correlations only. Although STGCN has a bigger RMSE than XGBoost on NYC Taxi, the outputs of STGCN show relatively high correlations on both two datasets. STG2Seq is a novel multi-step demand prediction architecture which combines long-term and short-term dependencies with attention mechanism. 
However, the zero padding for capturing spatial and temporal correlations simultaneously with simple graph convolution component might lead to the unsatisfactory results for STG2Seq.
Combining the sequence to sequence architecture for time series prediction with graph convolution contributes to the good performance of DCRNN. Benefit from adaptive adjacency matrix learning, Graph WaveNet also has competitive results. Our method achieves $13.85\%$ and $26.85\%$ RMSE lower than Graph WaveNet on two datasets, which indicates CCRNN captures transportation demand dependencies more accurately and efficiently.

\begin{figure}[bh!]
	\centering
	\includegraphics[width=1\columnwidth]{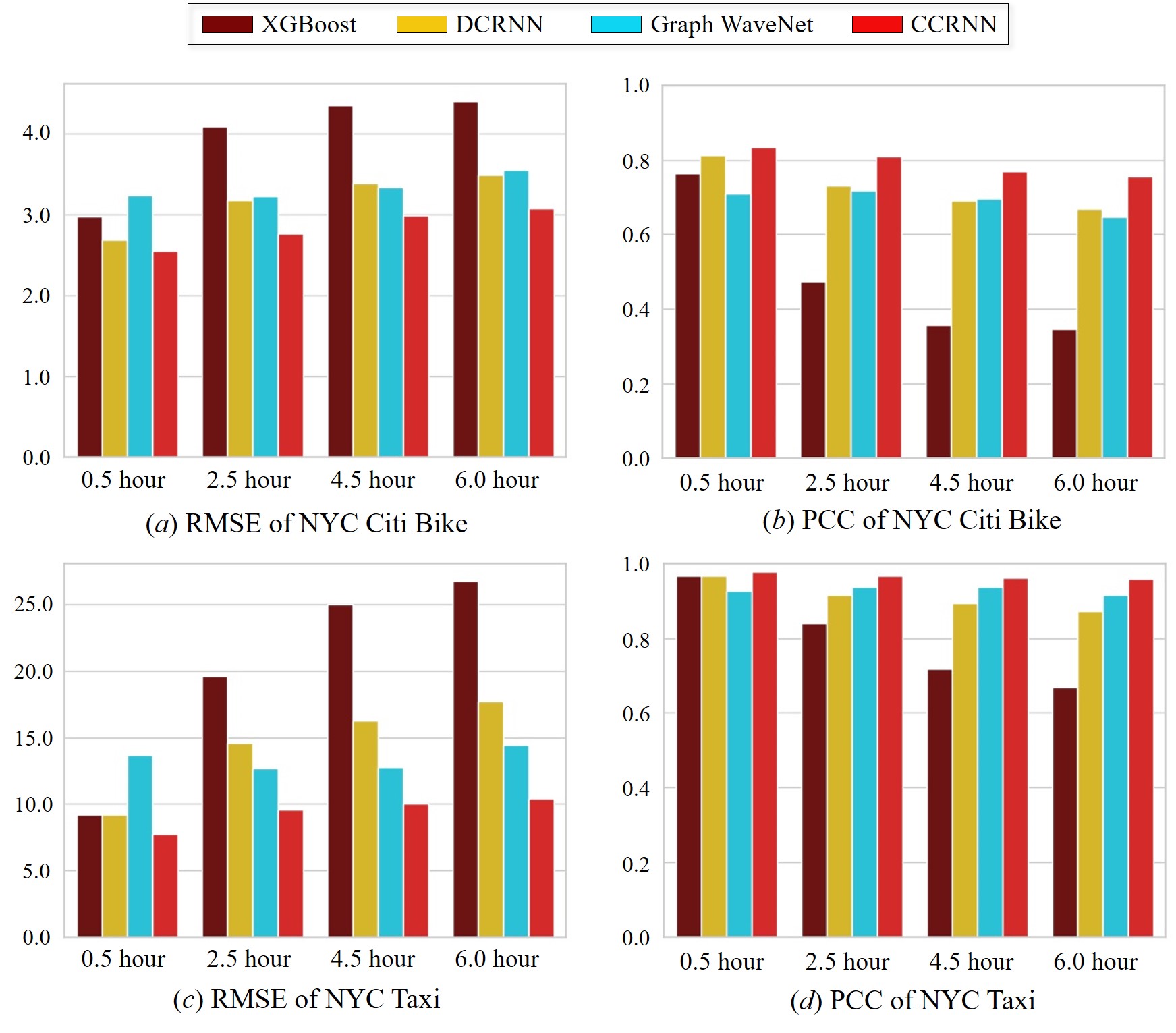}
	\caption{Comparison on different time steps.}
	\label{fig:experiment}
\end{figure}

\vspace{0.2 cm}
\noindent \textbf{Performances on Multi-step Demand Prediction.}
% \subsubsection{Performances on Different Steps.}
In the section, we illustrate the superiority of our model on single step prediction. Because of the space limitation, we evaluate our methods with several baselines at four specific time steps. 
% As it has shown in Figure \ref{fig:experiment}, we choose the first step (Step 1), the last step (Step 12), and two middle steps (Step 5 and Step 9) as the representatives. 
As it is shown in Figure \ref{fig:experiment}, we choose the first step (0.5 hour), the last step (6.0 hour), and two middle steps (2.5 hour and 4.5 hour).

XGBoost has the worst performance in middle-term and long-term predictions (2.5 hour, 4.5 hour and 6.0 hour) since its ignorance of spatial correlations.
Due to the recurrent prediction architecture, the performance of DCRNN declines as the forecast time becomes longer. Graph WaveNet outputs multi-step demands at once, which leads to stable results at different time steps, but relatively lower accuracy in short-term prediction (0.5 hour).
Although our model is in a sequence to sequence manner and suffers from the performance declining, we achieve the lowest RMSE and the highest PCC.

% \subsubsection{Ablation Study.}
% \vspace{0.25 cm}
% \noindent \textbf{Ablation Study.}
\subsection{Ablation Study}
We conduct an detailed ablation study on our method by removing or changing several components. In particular, five variants of our model are obtained by 1) No Adaptive: setting $\bm{E}_1^0$ and $\bm{E}_2^0$ untrainable, 2) No Coupling: removing the coupled mapping, 3) Random Init: randomly initializing $\bm{E}_1^0$ and $\bm{E}_2^0$, 4) Distance Init: initializing $\bm{A}^0$ with the distance between nodes \cite{li2017dcrnn}, 5) PCC Init: initializing $\bm{A}^0$ with PCC of demand time series \cite{bai2019stg2seq}.

As it is shown in Table \ref{tab:ablation study}, we could make a conclusion that the complete CCRNN achieves the best performance and the adjacency matrix generated by our method contains the most information. The adaptive and coupled layer-wise matrices help our model discover more accurate connectivity information and obtain high-level representations more efficiently.
Benefit from those components, although the variants initialized by the distance between stations and PCC of the time series obtain higher RMSE than CCRNN, their performances outperform all baselines in Table \ref{tab:baselinescomp}. 
Randomly initializing $\bm{E}_1^0$ and $\bm{E}_2^0$ leads to the poor performances with PCC lower than $0.1$. It is necessary for our method to get the adjacency matrix initialized properly to guide training. 

\begin{table}[tb!]
	\centering
	\caption{Comparison with variants of CCRNN.}
	\label{tab:ablation study}
	\begin{tabular}{c|c|ccc}
		\hline
		&Method & RMSE & MAE & PCC  \\
		\hline
		\multirow{6}{*}{\makecell[c]{NYC\\ Citi \\Bike}}&No Adaptive & 3.2648	&1.9157&	0.7143\\
		&No Coupling  & 3.0555 &1.8217 & 0.7702\\
		&Random Init. & 4.7308& 3.0366 & 0.0204\\
		&Distance Init. & 2.9718 & 1.7995 & 0.7768\\
		&PCC Init. &  2.9096& 	1.7729& 0.7851 \\
		&\textbf{CCRNN} & \textbf{2.8382} & \textbf{1.7404} & \textbf{0.7934}  \\
		\hline
		\multirow{6}{*}{\makecell[c]{NYC\\ Taxi}}&No Adaptive & 12.0516	&7.2842	&0.9404\\
		&No Coupling   & 10.1864&	5.8168	&0.9594\\
		&Random Init. & 36.0307& 22.2210 & 0.0065\\
		&Distance Init. & 10.6882&	6.1476&0.9559\\
		&PCC Init. &  10.0244&	5.7927&	0.9619\\ 
		&\textbf{CCRNN} &\textbf{9.5631} &\textbf{5.4979} &  \textbf{0.9648}\\
		\hline
	\end{tabular}
\end{table}

\section{Conclusion}

% In this paper, we presented a novel transportation demand prediction model, namely, xxxx (xxx), and a data-driven adjacency matrix generation method. In an encoder-decoder forecasting architecture, we hierarchically aggregated the multi-level representations extracted by Evolution Graph Convolution (EGC). 
% We proposed an external factors free adjacency matrix generation method which captures the temporal patterns' similarity.
% Experiments were conducted on real-world taxi and sharing bike datasets, and state-of-the-art results were achieved by ECRNN. 
% This research represented a new perspective in graph convolutional network with evolutionary matrices. In the future, we will explore performances of xxx on other graph convolution tasks.

% In this paper, we presented a novel transportation demand prediction model, namely, CCRNN. Specifically, to address the problem that it is difficult for the popular graph convolutional networks to capture multi-level spatial dependence efficiently and accurately, we proposed a novel graph convolution architecture, CGC, with self-learned adjacency matrices varying from layer to layer. Furthermore, by modeling the layer-wise topological structure correlations, we provided a coupled mapping mechanism to realize this graph convolution structure in a low computational cost manner. The extracted representations were attached different importance by a multi-level aggregation. Above components were fused with a recurrent unit to capture spatio-temporal correlation simultaneously.

In this paper, we presented a novel transportation demand prediction model, namely, CCRNN. In particular, to capture multi-level spatial dependence, we proposed a novel graph convolution architecture, CGC. The adjacency matrices in CGC were self-learned and varied from layer to layer. Furthermore, a layer-wise coupling mechanism was employed to bridge the upper-level graph structure with the lower-level one. It also reduced the scale of parameters in our model. Then, the different importance was attached to extracted representations by a multi-level aggregation module.
A unitary network fused the above components to make final predictions. 
% The extracted representations were attached different importance by a multi-level aggregation module. Above components were fused with a recurrent unit to capture spatio-temporal correlation simultaneously.
% To capture multi-level spatial representations, we proposed a novel graph convolution architecture with adjacency matrices varying from layer to layer. The adjacency matrices discovered the genuine hierarchical proximity of graph structure in the self-learning process.
% The graphs in upper layer were built upon on lower layer topological structure. In a sequence to sequence prediction architecture, temporal correlations were capturing simultaneously by integrating above components with a GRU unite. 
Experiments were conducted on real-world taxi and sharing bike datasets, and state-of-the-art results were achieved by CCRNN.
This research represented a new perspective in graph convolutional network with layer-wise adjacency matrices. In the future, we will study performances of CGC on other graph convolution tasks.

\section{Acknowledgments}
% We thank the anonymous reviewers for their constructive comments on this research work.
This work is supported by the National Natural Science Foundation of China under Grant No. 51778033, 51822802, 51991395, 71901011, U1811463, the National Key R\&D Program of China No. 2018YFB2101003, the Science and Technology Major Project of Beijing under Grant No. Z191100002519012.
\clearpage

\bibliography{Reference}
\end{document}